\documentclass[preprint,12pt]{elsarticle}

\usepackage{graphicx}
\usepackage[cmex10]{amsmath}
\usepackage{amsthm}
\usepackage{multirow}
\usepackage{color}
\usepackage{hyperref}
\usepackage{amssymb}
\usepackage{import}
\usepackage{hyperref}
\usepackage{flushend}
\usepackage{balance}
\usepackage{booktabs}
\usepackage[T1]{fontenc}
\usepackage[font=small,labelfont=bf]{caption}
\usepackage[table]{xcolor}
\usepackage{array}
\usepackage{url}
\usepackage{natbib}
\usepackage{lineno}
\providecommand{\bet}[1]{{{\colorbox{gray}{\textcolor{white}{\textbf{#1}}}}}}

\newenvironment{myenum}{
\begin{enumerate}
 \setlength{\itemsep}{1pt}
 \setlength{\parskip}{0pt}
 \setlength{\parsep}{0pt}
}{\end{enumerate}}

\usepackage[]{algorithm2e}

\providecommand{\bet}[1]{{{\colorbox{gray}{\textcolor{white}{\textbf{#1}}}}}}

\providecommand{\betPruning}[1]{{{\colorbox{gray!40}{\textcolor{black}{\textbf{#1}}}}}}
\usepackage{comment}
\usepackage{multirow}
\usepackage{multicol}
\usepackage{paralist}
\usepackage{tikz}
\usepackage{longtable} 
\usepackage{indentfirst}
\usepackage{multirow}
\usepackage{soul}
\usepackage{adjustbox}
\usepackage{booktabs}
\usepackage{adjustbox}
\usepackage{tikz}
\usepackage{subfig}
\usepackage{xcolor,colortbl}
\usepackage{diagbox}
\usepackage{pdflscape}
\usepackage{ulem}

\begin{document}

\begin{frontmatter}

\title{An Evolutionary Approach for Creating of Diverse Classifier Ensembles}

\author{\'{A}lvaro R. Ferreira Jr, Fabio A. Faria, Gustavo Carneiro, and Vin\'{i}cius V. de Melo}
\address{Institute of Science and Technology\\ Universidade Federal de S\~{a}o Paulo (ICT-Unifesp)\\ aalvin10@gmail.com, dr.vmelo@gmail.com, ffaria@unifesp.br}
\address{Australian Institute for Machine Learning \\ The Universidade of Adelaide \\gustavo.carneiro@adelaide.edu.au }

\begin{abstract}
Classification is one of the most studied tasks in data mining and machine learning areas and many works in the literature have been presented to solve classification problems
for multiple fields of knowledge such as medicine, biology, security, and remote sensing. The objectives of classification are to find a hypothesis (model) that fits a training set and generalizes to unseen test sets. Since there is no single classifier that achieves the best results for all kinds of applications, a good alternative is to adopt classifier fusion strategies. One type of classification fusion relies on an ensemble of classifiers that aims to combine different and complementary classifiers to improve the prediction accuracy on a test set. A fundamental concept in classification ensemble is the measure of diversity between classifiers. Although many authors have adopted the use of diversity in classification ensemble, it has been found that its use alone is not sufficient to achieve effective improvements over the best-performing individual classifiers. A key point in the success of these approaches is the combination of diversity and accuracy among classifiers belonging to an ensemble. With a large amount of classification models available in the literature, one challenge is the choice of the most suitable classifiers to compose the 
final classification system, which generates the need of classifier selection strategies. We address this point by proposing a framework for classifier selection and fusion based on a four-step protocol called $CIF-E$ (\underline{\textbf{C}}lassifiers, \underline{\textbf{I}}nitialization, \underline{\textbf{F}}itness function, and \underline{\textbf{E}}volutionary algorithm). Our approach uses evolutionary algorithms to combine the most diverse base classifiers from a set of available classifiers into a final ensemble to optimize the prediction accuracy. We implement and evaluate 24 varied ensemble approaches following the proposed $CIF-E$ protocol and we are able to find the most accurate approach. A comparative analysis has also been performed among the best approaches and many other baselines from the literature. The experiments show that the proposed evolutionary approach based on Univariate Marginal Distribution Algorithm (UMDA) can outperform the state-of-the-art literature approaches in many well-known UCI datasets.
\end{abstract}

\begin{keyword}
diversity measure \sep ensemble of classifiers \sep ensemble selection.
\end{keyword}

\end{frontmatter}

\section{Introduction}

Classification is one of the most studied tasks in machine learning. 
Much research has been done in the development of classifiers to solve several problems in different knowledge domains (e.g., medicine~\cite{medicine}, biology~\cite{biology}, agriculture~\cite{agriculture}, security~\cite{security}, and remote sensing~\cite{remotesensing}). To perform a classification task, one needs to build a classifier, which consists of finding a hypothesis/model that properly fits the training data distribution and generalises well in the test set. One of the main challenges in this process is to find the classifier that will produce the best possible fitting and generalisation results for a particular classification problem. 

According to Wolpert~\cite{wolpert_1996}, there is no single technique that can solve all problems in different application domains ("{No Free Lunch}" Theorem). One possible solution to mitigate such issue is the employment of classification fusion, also called ensemble of classifiers~\cite{WOZNIAK20143}. Ensembles of classifiers are learning algorithms that gather different classifiers to classify new data by taking a (weighted) vote of their predictions. Dietterich~\cite{Dietterich2000} showed the benefits of using those algorithms focusing on three fundamental issues: statistical, computational, and representational.

Due to the inability to use all classification techniques available in the literature, a challenge lies in being able to select a small subset of classifiers from a pool, where this subset can achieve an ensemble generalization performance (prediction quality) better than the ensemble consisting from the initial pool of classifiers. Such a subset of classifiers requires less storage and computational resources, and thus, improves the final prediction efficiency~\cite{Zhou2012}.

According to Tumer and Ghosh~\cite{Tumer96}, a  crucial concept to yield good performance with an ensemble of classifiers is the use of diversity. Diversity assesses the degree of agreement/disagreement among the classifiers in a pool, which can help identify potential classifiers to build the ensemble. One can produce diversity in many ways~\cite{Zhou2012}:
(1) Data Sample Manipulation generates multiple training sets (e.g., the bootstrap sampling on the Bagging technique~\cite{bagging1996});  (2) Input Feature Manipulation selects different subsets of features (sub-spaces) to provide many views of the data (e.g., Random Forest approach~\cite{randomforest2001}); (3) Learning Parameter Manipulation tries to create diverse classifiers by using different parameter settings for the learning method (e.g., initial weights assigned to neural networks~\cite{Kolen:1990});
(4) Output Representation Manipulation combines many classifier outcomes to create a final prediction (e.g., ECOC approach~\cite{ecoc1994}). 

However, analyzing only the diversity concept among classifiers has limitations. Kuncheva~\cite{Kuncheva03thatelusive} argued that although diversity is important in classifier ensembles, it is an elusive concept and does not guarantee good results.  Zhi-Hua Zhou~\cite{Zhou2012} mentions that the combination of poor and diverse classifiers does not improve performance and can even worsen the effectiveness of the ensemble. Hence, although diverse is important, one must not forget to take into account classification accuracy.

In literature, many works have adopted GA techniques for optimization problems in different application domain (e.g., biology~\cite{valikhan2019development}, engineering~\cite{yu2015application}). However, as disadvantages to use those techniques are: (1) the premature convergence usually caused by loss of diversity within the population~\cite{malik2014preventing}; and (2) high  sensitivity to initial population~\cite{valikhan2019development}. 
On the opposite way, the Estimation  of  Distribution  Algorithms (EDAs) avoid premature convergence as well as use of a compact and short representation solve many large and complex problems in different applications domain~\cite{UMDA_2011,zhang2017association,utamima2019evolutionary,eda_2011}.


In this sense, this paper proposes the development of a new framework for classifier selection and fusion, combining diversity measures and accuracy through evolutionary algorithms (GA and EDA) to select the most suitable base classifiers from a pool of classifiers. Our method enables the creation of an ensemble of classifiers to improve the classification accuracy when compared with the results of individual classifiers. Therefore, the main contributions of this paper are:
\begin{myenum}
  \item A new flexible framework for classifier selection and fusion based on evolutionary algorithms;
  \item A new approach for the creation of an initial population in evolutionary algorithms based on the ranking aggregation technique; 
\end{myenum}

This paper is organized as follows: Section~\ref{sec:relwork} explains important concepts and related works for a better understanding of this paper. Section~\ref{sec:framework} shows the framework for classifier selection and fusion based on evolutionary algorithms in a four-step protocol named $CIF-E$. Section~\ref{sec:methodology} shows the experimental methodology adopted in this work. Section~\ref{sec:experiments} discusses the experimental results using $CIF-E$ protocol. In Section~\ref{sec:conclusion}, some conclusions and future work are presented. 

\section{Related Work}
\label{sec:relwork}

This section explains some important concepts and related works necessary for a better understanding of our paper.

\subsection{Evolutionary Algorithms}
We present the two different evolutionary algorithms adopted in this work.

\subsubsection{Genetic Algorithm}
Holland \cite{holland1992adaptation} introduced the Genetic Algorithm (GA), where the goal was to study the phenomenon of adaptation as it occurs with animals in the wild and develop techniques in which natural adaptation mechanisms could be employed in optimisation problems. The GA is a method for evolving populations of individuals or "chromosomes", usually represented by bit strings, to a newer, better fit population, where the fitness is related to an individual's effectiveness in solving specific problems. Using natural selection approaches with genetically inspired crossover and mutation operators, the selection operator chooses the chromosomes in the population that will be allowed to reproduce, where the fittest individual is more likely to be chosen. 

The simplest GA has three core operators: selection, single-point crossover, and mutation. Selection is the factor that guides the algorithm to the solution, giving preference to individuals with high fitness, commonly used with techniques that exploit randomness (e.g., roulette selection). A simple pseudo-code is presented in Algorithm 1 and explained next.


To generate fitness proportionate selection (roulette), one uses the fitness value of each individual: that $N$ individuals will be allocated in a roulette, the probability of selecting individuals is related to their fitness, i.e., the best fit individual has the largest probability (area in the roulette) while the worst fit has the lowest probability. To choose an individual, the roulette is ``rotated'' and the individual corresponding to the selected roulette position is chosen.

In the crossover, the chromosomes of two parents (represented as $Ind_{1}$ and $Ind_{2}$) are combined interchangeably. In the simplest way, this process is performed with a single cutoff point $X$, which is randomly chosen. The child is then formed by the genes from $Ind_{1}$ before the point $X$ concatenated with the genes from $Ind_{2}$ starting at $X$. Such genetic operator is responsible for making big changes in the solution, resulting in big jumps in the search-space. However, this operator may reduce the difference between children and parent leading to a loss of population diversity. To help with this issue, one needs another operator.

Mutation is a method of changing an individual's genes in a similar way as it happens in nature. This operator acts by randomly selecting and changing one or more genes in the solution generated by the crossover operator (the child solution). This method requires of probability to be applied to genes, thus a low value will result in mutating only a few genes while a large value may mutate the entire individual. While this operator is good for introducing diversity, it might also worsen the optimization because it can avoid convergence.



In order to reduce the randomness present in the GA, other algorithms have been developed to perform a better search -- we present some of these algorithms below.

\subsubsection{Univariate Marginal Distribution Algorithm (UMDA)}
Introduced by Mühlenbein \& Paa\ss~\cite{muhlenbein1996recombination}, the {Univariate Marginal Distribution} algorithm (UMDA) is one of the simplest Estimation of Distribution Algorithms~\cite{HAUSCHILD2011111} (EDAs). To optimize a pseudo-Boolean function $f: \{0,1\}^n \rightarrow \mathbb{R}$ where an individual is a bit-string (each gene is 0 or 1), the algorithm follows an iterative process: 1) independently and identically sampling a population of $\lambda$ individuals (solutions) from the current probabilistic model; 2) evaluating the solutions; 3) updating the model from the fittest $\mu$ solutions. Each sampling-update cycle is called a generation or iteration. In each iteration, the probabilistic model in the generation $t \in \mathbb{N}$ is represented as a vector $p_t=(p_t(1),...,p_t(n))\in [0,1]^n$, where each component (or marginal) $p_t(i)\in[0,1]$ to $i\in[n]$, and $t \in \mathbb{N}$ is the probability of sampling the number one in the $i^{th}$ position of an individual in generation $t$. Each individual $x = (x_1, ..., x_n) \in \{0, 1\}^n$ is therefore sampled from the joint probability.

\begin{equation}
    Pr(x | p_t) = \prod_{i=1}^{n}p_t(i)^{x_i}(1-p_t(i))^{(1-{x_i})}.
    \label{equacaoPr}
\end{equation}

   

\begin{figure}[ht!]
\centering
\includegraphics[width=0.7\textwidth]{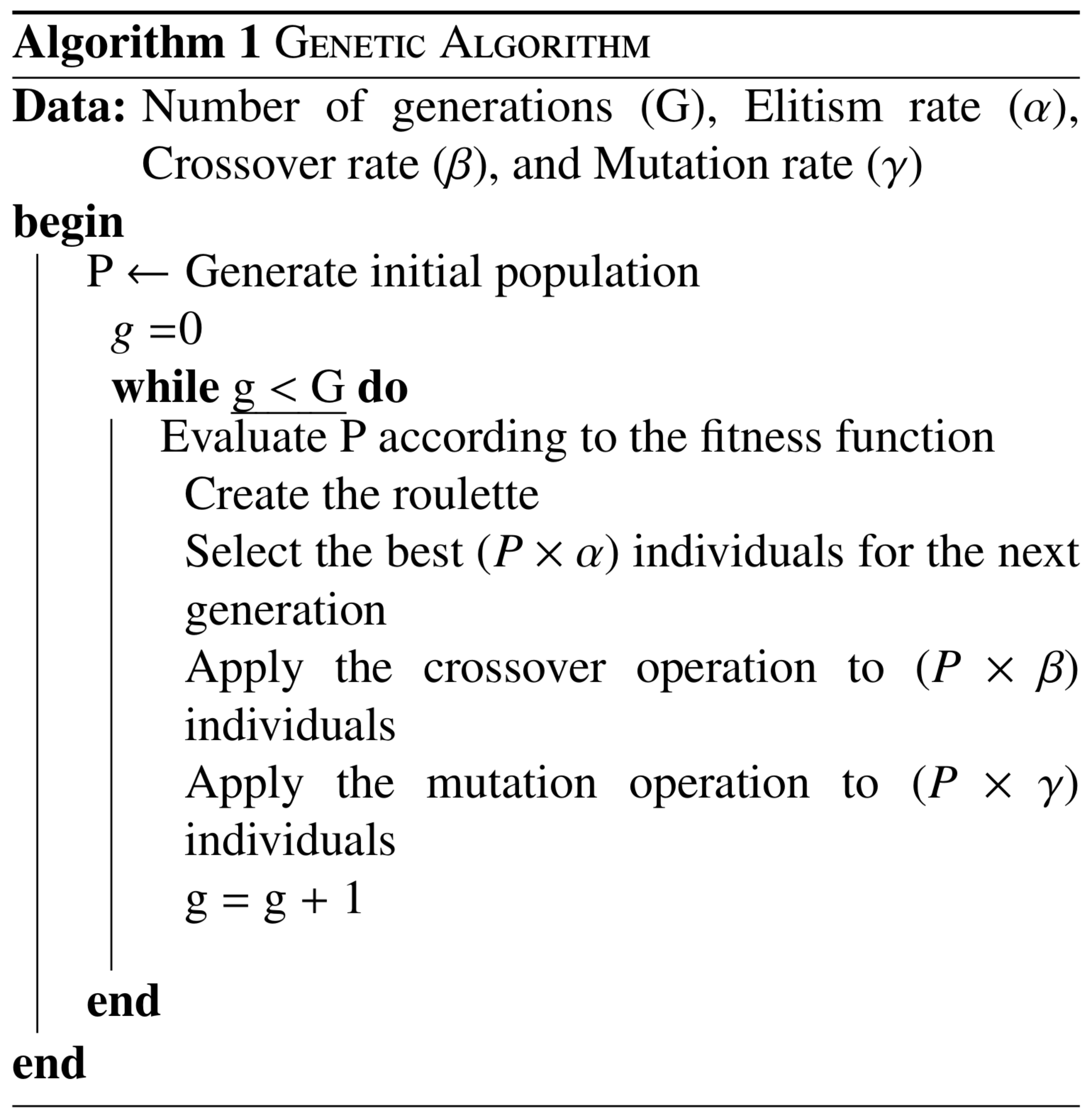}
\label{estruturaAlgoritmoGenetico}
\end{figure}



Extreme probabilities, zero and one, should be avoided for each marginal point $(i)$; otherwise, the bit at position $i$ would never have its value changed, making the optimization ignore some regions of the search space. To avoid this, all $p_{t + 1} (i)$ margins are usually restricted within the closed range $[1/n, 1-1/n]$ and these values $1/n$ and $1-1/n$ are called lower and upper bounds respectively. In this case, UMDA is known as "UMDA with margins".

\begin{figure}[ht!]
\centering
\includegraphics[width=0.7\textwidth]{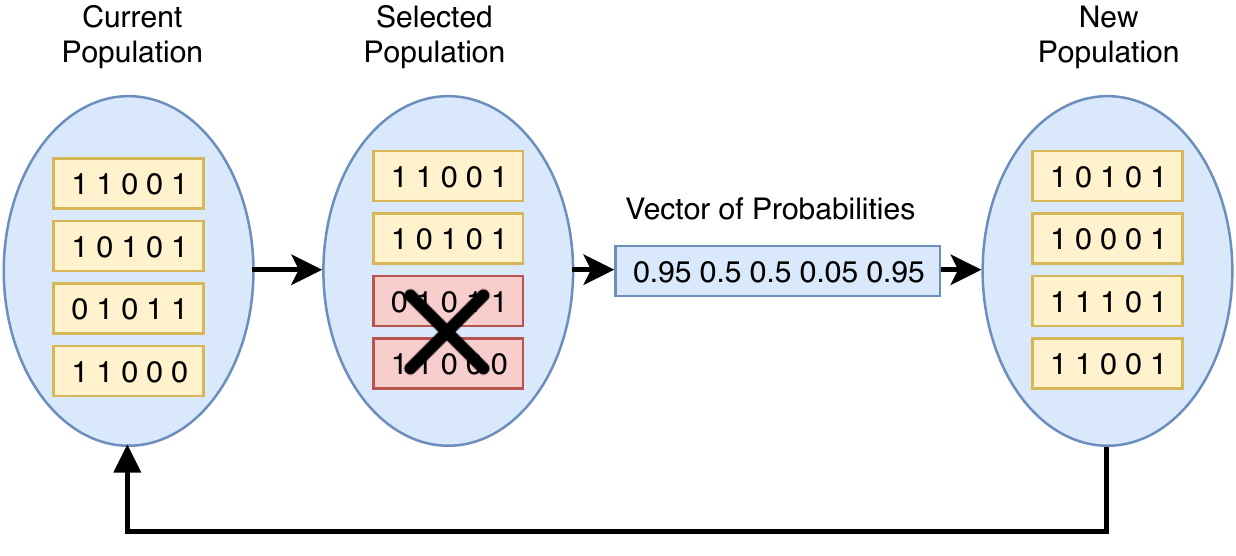}
\caption{UMDA algorithm with margins, i.e., lower and upper bounds of $5\%$ and $95\%$, respectively. Extracted and adapted from~\cite{martinpelikan;kumarasastry2009}.}
 \label{fig:umdaprocessodescritivo}
\end{figure}

Figure~\ref{fig:umdaprocessodescritivo} illustrates the UMDA with margins. Suppose the current population has four individuals composed of five genes, which are evaluated and ranked by a fitness function. Let $50\%$ of the population ($\mu=0.5$) be selected and the genes probabilities are recalculated $(0.95, 0.5, 0.5, 0.05, 0.95)$ as the probability each gene is set as one. It is important to note that although the genes in positions $1$ and $5$ appear in $100\%$ of the individuals on the selected population, their probabilities are $95\%$ and for gene $4$ it is $5\%$ because of the margins explained before.

Algorithm 2 shows the pseudocode of the UMDA with margins.
\begin{figure}[ht!]
\centering
\includegraphics[width=0.7\textwidth]{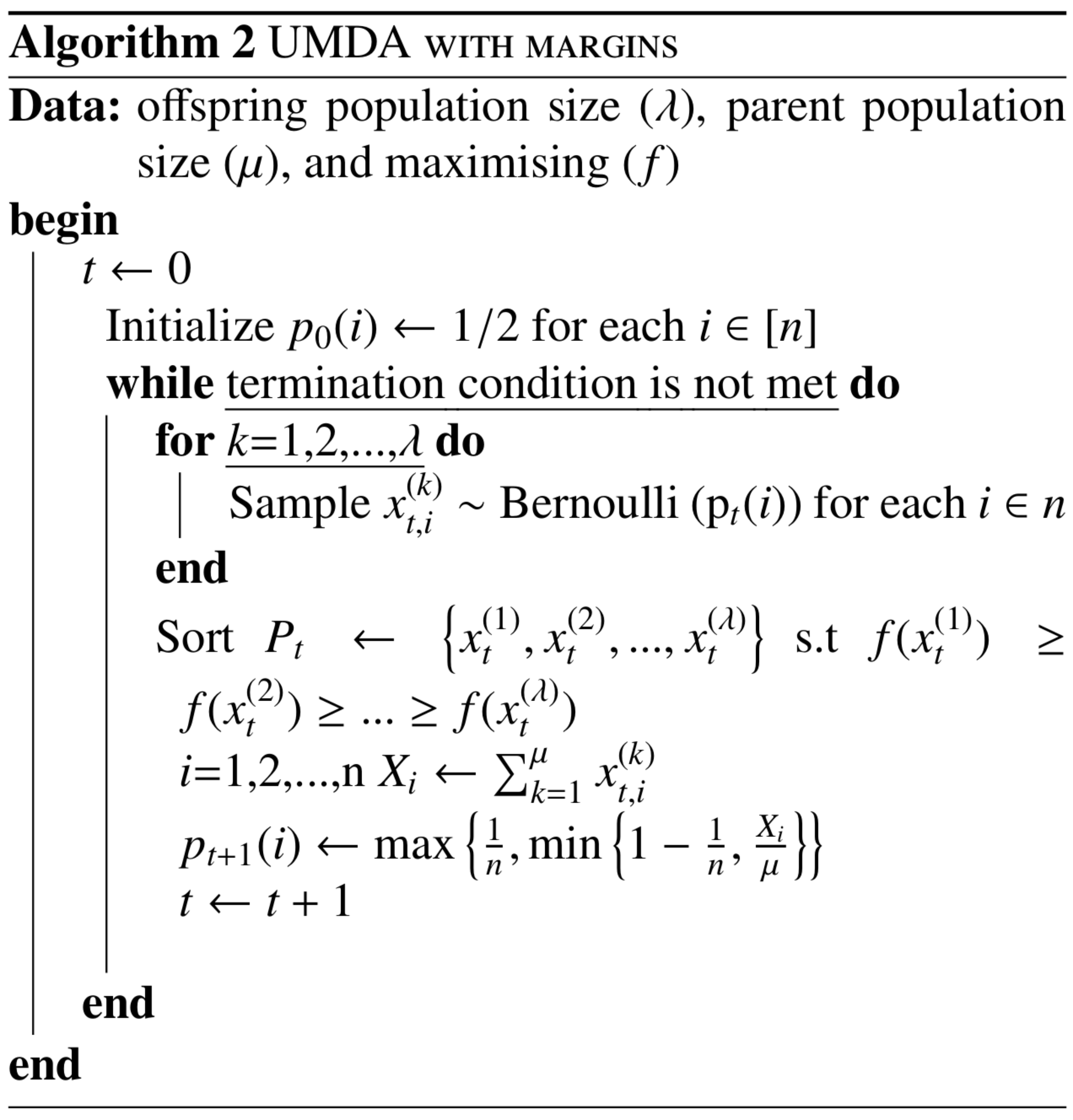}
\label{pseudocodigoumda}
\end{figure}


 
Therefore, the search is based on sampling models of the variables, not randomly crossing and mutating them. Although the models treat the variables as i.i.d., this approach leads to a better optimization process
by reducing the generation of poor-quality solutions and increasing the generation of better-quality solutions.

\subsection{Ensemble Selection Approaches}
This section shows different ensemble selection approaches used as baselines in this work.

\subsubsection{Aggregation Ordering in Bagging (AGOB)}
AGOB is a strategy for improving bagging generalization performance by selecting subsets of classifiers ($C^* \in C$). In this approach, the goal is to find a good classifiers aggregation order in the bagging technique. The new ordering can be obtained by different rules such as reduced-error pruning, a complementary measure, and margin distance minimization.  
Finally, the first $C^*$ classifiers are chosen depending on the desired amount of pruning: given an initial subset of size $u-1$, a subset of size $u$ is obtained by adding a classifier that is selected by an appropriate evaluation measure~\cite{AGOB_2004}.

\subsubsection{Genetic Algorithm-based Selective ENsemble (GASEN)}

GASEN (Genetic Algorithm-based Selective ENsemble)~\cite{GASEN_2002} is a pruning approach that trains $N$ individual neural networks and then employs a GA to select an optimal subset of individuals with the objective of increasing the generalization quality. Each individual neural network has a weight $w_i; i=1,...,N$, where $0<w_i<1$ and $\sum_{i=1}^{N} w_i = 1$, that measures its importance in the set; the networks where $w_i$ is greater than a predefined threshold $\lambda$ are selected to form the subset of classifiers. Since the ideal weight must minimize the generalization error of the set, considering Equation~\ref{eq:equacao3gasen}, the optimal weight of the weight vector $ w_ {opt} $ can be expressed as:

\begin{equation}
    w_{opt} = argmin\left ( \sum_{i=1}^{N}\sum_{j=1}^{N}w_{i}w_{j}C_{ij} \right ), i<>j
    \label{eq:equacao3gasen}
\end{equation}

\noindent where $C_{ij}$ corresponds to the correlation between the $i^{th}$ and the $j^{th}$ neural networks.

GASEN uses GA to optimize the weight vector and discover the appropriate subset of individual networks.
According to the authors, the use of the weight vector in the selection of the individual networks and in the combination of the individual predictions can easily to cause overfitting. Therefore, the weights are not used to calculate the output of the ensemble.

\subsubsection{ Combining Diversity Measures for Ensemble Pruning (DivP)}

This approach is a method that combines several diversity measures (disagreement, Q statistic, correlation coefficient, Kappa statistic, and double failure measure) to {prune} an original set of classifiers into subsets of classifiers consisting of two main components: 1) diversity, based on the initial population of classifiers and 2) the pruning method which is performed with the aid of GA. The set of classifiers $C$ is created using bagging strategy, where the base classifiers are represented by Perceptrons trained with bootstrapped data to produce significant diversity. 

The output of the diversity measure module consists of a $|C| \times |C|$ matrix of weights that show how to combine the classifiers, where the weights are optimized by GA. The chromosome of the GA consists of six values: weights $w_i, i=1,...,5$ to combine the diversity matrices and the threshold $T$ to calculate the adjacency matrix. 

After the matrix generation, the candidate set is constructed from the adjacency matrix, where the classifiers with a high degree of diversity are put together. To perform such operation, a graph coloring algorithm was employed so that the adjacent vertices have different colors. Candidate sets are created by grouping vertices (classifiers) with the same color; then, the best candidate set that minimizes the error rate in the validation set is selected~\cite{DIVP_2016}.

\subsubsection{Diversity Regularized Ensemble Pruning (DREP)}
DREP initializes a set with the best classifier obtained from the validation set and iteratively inserts new classifiers into this set based on error rate and diversity. At each step, candidate classifiers are evaluated according to their differences with the current set, and the classifier which reduces the error rate the most and increases diversity is selected to compose the set. The selection criteria for choosing a classifier is weighted by the $\rho$ parameter, where high values mean more emphasis on error and less emphasis on diversity, and vice-versa for low values of $\rho$~\cite{DREP_2012}.

\subsubsection{Pruning Adaptive Boosting (Kappa)}
In the Kappa \cite{kappa_pruning_1997} approach, the diversity measure {kappa} is calculated for all classifier pairs generated using the {AdaBoost} method. Thus, iteratively, the classifier pairs with the lowest value of $\kappa$ are selected for the final set until they reach a predefined {threshold}.
To get the value of $\kappa$, given two classifiers $h_{\alpha} $ and $h_{\beta}$, and a dataset containing $m$ examples, a contingency table $CT$ is constructed where cell $CT_{ij}$ contains the number of $x$ examples for which $h_{\alpha}(x)=i$ and $h_{\beta}(x)=j$. If $h_{\alpha}$ and $h_{\beta}$ are identical in the dataset, then all nonzero counts will appear diagonally. If $h_{\alpha}$ and $h_{\beta}$ are very different, there will be large off-diagonal values. Let

\begin{equation}
    \Theta_1 = \frac{\sum_{i=1}^{L}CT{_{ii}} }{m}
\end{equation}

\noindent be the probability that two classifiers agree, then we use $\Theta_1$ as a measure of agreement. On the other hand, $\Theta_2$ can be defined as the probability that two classifiers agree by chance, given the values observed in the contingency table, $ \Theta_2$ is defined by.

\begin{equation}
    \Theta_2 = \sum_{i=1}^{L}\left ( \sum_{j=1}^{L} \frac{CT{_{ij}}}{m}\cdot \sum_{j=1}^{L} \frac{CT{_{ji}}}{m} \right ).
\end{equation}
\noindent

Therefore,the value of $\kappa$ is defined as:

\begin{equation}
    \kappa = \frac{\Theta_1 - \Theta_2}{1 - \Theta_2},
\end{equation}
\noindent

where $\kappa = 0$ if the concordance of two classifier is equal to a random concordance and $\kappa = 1$ if both classifiers agree on each sample.

\subsubsection{Pruning in Ordered Bagging Ensembles (POBE)}

In POBE, the process is based on sorting the bagged classifiers and then selecting subsets to be combined. Sorting the sets allows us to initially select the best classifiers and add classifiers gradually through some evaluation criterion. 
The evaluation of each subset consists of counting how many classifiers correctly predicted on each validation example. The effectiveness of the subset is the sum of correct predictions regarding the entire validation set. For each subset approaching the optimal result, a new classifier is inserted that maximizes the generalization of the subset. The process continues until the subsets have between $15\%$ and $30\%$ of the initial set of classifiers~\cite{POBE_2006}.

\section{A Framework for Classifier Selection and Fusion based on Evolutionary Algorithms}
\label{sec:framework}

In this section, we describe in details the proposed framework for classifier selection and fusion.

\subsection{Overview} \label{subsec:overview}

\begin{figure}[ht!]
\centering
\includegraphics[width=0.7\textwidth]{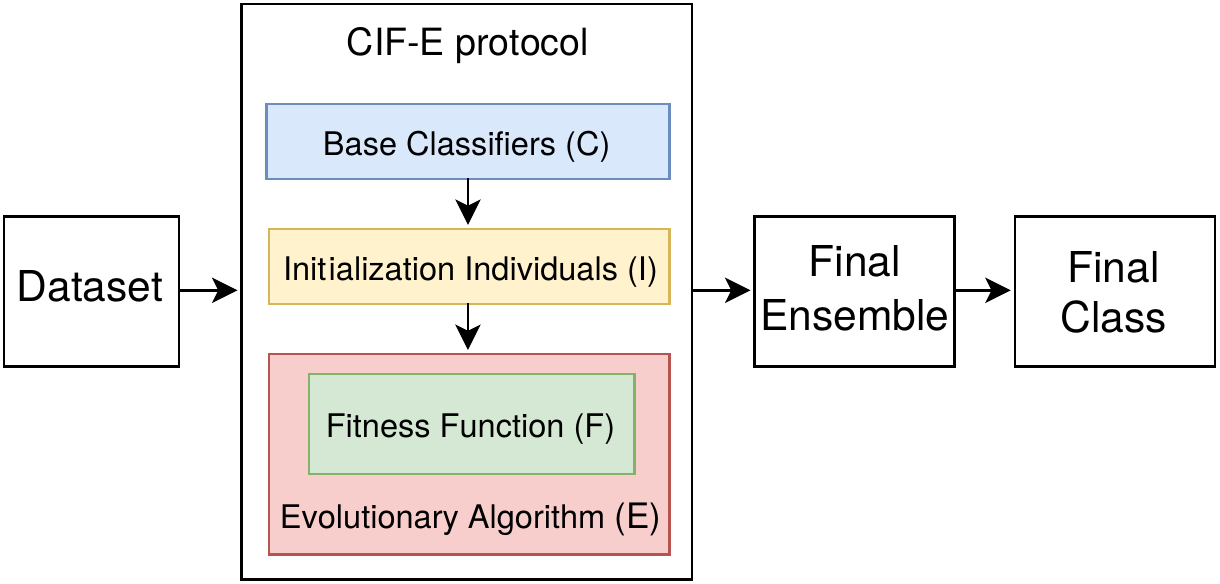}
\caption{A pipeline of the proposed framework for classifier selection and fusion based on an evolutionary algorithm. The colored parts represent the $CIF$-$E$ protocol adopted in this work.}
\label{fig:arcaboucoBR}
\end{figure}

Figure~\ref{fig:arcaboucoBR} illustrates the proposed framework for classifier selection and fusion, which is composed of a four-step protocol called $CIF$-$E$ (\underline{\textbf{C}}lassifiers, \underline{\textbf{I}}nitialization, \underline{\textbf{F}}itness function, and \underline{\textbf{E}}volutionary algorithm). 

Initially, there is a process of selecting the type of the machine learning technique which will be used to generate the base classifiers of the ensembles. This first step represented by \underline{\textbf{C}} can contain two modes of operation: 1) (P) for approaches that use only one type of technique or 2) (M) for ones that use different learning techniques. 

In the second step, the strategy of individual initialization (population), represented by \underline{\textbf{I}}, can be performed randomly (A) or by an approach named here as \textit{tuning}~(T). 

In the third step, all generated individuals are evaluated by a fitness function represented by \underline{\textbf{F}} that can be of three modes: 1) (E) is a fitness function that considers only the error rate of the individual/ensemble; 2) (D) is a fitness function that uses the error rate and the diversity score existing in the ensemble; or 3) (P) is a fitness function that considers the two previous approaches (error and diversity) and includes the amount of classifiers present in the ensemble. 

Finally, in the fourth step, represented by \underline{\textbf{E}} in the $CIF$-$E$ protocol, the optimization of the solutions is accomplished through evolutionary algorithms such as Genetic Algorithm (GA) and Univariate Marginal Distribution Algorithm (UMDA).

\subsection{Strategies for Selecting Base Classifiers (C)}

This step consists in defining what machine learning techniques will be used in the generation of the base classifiers, which may be one type of technique (P) or multiple techniques (M). In addition, different classifiers will be created by a re-sampling strategy in the training set using the bootstrap process to increase the diversity among the base classifiers present in the pool. To prevent low-performing base classifiers from entering into the pool, $3,000$ base classifiers are initially generated and only the best $|\mathcal{C}|=\{50,100,150,200,250\}$ of them are selected for the initial pool.

\subsection{Strategies for Choosing Initial Individuals (I)}

After performing the generation of base classifiers, the next step in the framework is the generation of solutions (individuals). An individual can be represented as a binary vector of size $|\mathcal{C}|$, where $C_i=1$ indicates the $i^{th}$ classifier 
is active in the ensemble. 

The generation of individuals (population) can be performed randomly (A) or through a strategy based on ranking aggregation developed in this work called \textit{tuning} (T).

\subsubsection{Random (A)}
The random initialization process occurs without considering any information about the available base classifiers, i.e., the classifiers are randomly selected to compose the ensembles. The number of classifiers to be inserted into each ensembles is also randomly determined.

\subsubsection{{\textit{Tuning}} (T)}
\label{sec:tuning}

The \textit{Tuning} initialization is a process of creating the initial population based on the ranking aggregation technique. This process is described in details below.

\begin{enumerate}[(i)]

\item \textbf{Computing Diversity Measure Matrices:} 

After all base classifiers are generated, each one predicts all instances on the validation set (validation $1$) to create a label matrix. Such matrix is used to compute $D$ pairwise diversity measures among all pairs of classifiers in the pool. Each diversity measure gives rise to a ranked list with ${\frac{|\mathcal{C}| \times (|\mathcal{C}|-1) }{2}}$ rows. 

Let $\mathcal{R}_{d_l}=$ \{$(c_{i}, c_{j})$ , $score_{d_{l}}(c_{i},c_{j})$\} be a ranked list of pairs of classifiers defined by the score of the diversity measure $d_{l} \in {D}$ of pairs of classifiers $(c_{i}, c_{j})$. Consider that low values of $score_{d_{l}}(c_{i}, c_{j})$ indicate high diversity between the pair $(c_{i}, c_{j})$, and therefore, the most suitable pairs of classifiers to be combined are at the top positions of ranked list $\mathcal{R}_{d_l}$.  In case of the diversity measure for which high score values indicate high diversity, we used the inverse of this measure (e.g., Disagreement Measure).

\item \textbf{Diversity-Accuracy Rank Aggregation:} 

Let ${R}_{d}$ = \{${R}_{d_1}, {R}_{d_2} \ldots {R}_{d_{|{D}|}}$\} be a set of ranked lists ($l$). The  $\textit{score}_{d_{c}}$ is computed through the multiplication of each $score_{d_l}$ from the ranked lists ${R_{d_l}} \in R_d$:

\begin{equation}
    \label{scoreunico}
    \textit{score}_{d_{c}} (c_{i}, c_{j}) = \prod_{l = 1}^{|{D}|} (1+score_{d_{l}} (c_{i}, c_{j})).
\end{equation}

We employed Equation~\ref{scorerank} for creating the final ranked list ${R}_{rank}$, which orders the pairs of classifiers with high diversity rate and high accuracy rate at the top positions:

\begin{equation}
\label{scorerank}
    \textit{score}_{rank} (c_{i}, c_{j}) = \alpha \times error(c_{i}, c_{j}) + (1-\alpha) \times \textit{score}_{d_{c}} (c_{i}, c_{j}).
\end{equation}

Let $\alpha \in (0,1)$ and ($1$-$\alpha$) be the weighting coefficients for diversity score ($score_{d_{c}}$) and error rate ($error(c_{i},c_{j})$) in validation set (in this work, $\alpha=0.5$). The ${error(c_{i},c_{j}) = 1 - accuracy(c_{i},c_{j})}$, where $accuracy(c_{i},c_{j})$ is the mean accuracy between the  classifiers $c_{i}$ and $c_{j}$. Thus, the most suitable pairs of classifiers will be at the top of the final ranked list positions defined in Equation~\ref{eq:listrank}. 

\begin{equation}
\label{eq:listrank}
    \mathcal{R}_{rank} = \{(c_{i},c_{j}), score_{rank}(c_{i},c_{j})\}.
\end{equation}

Note that these pairs of classifiers minimize $score_{rank}(c_{i},c_ {j})$. 

\item \textbf{Creating a Histogram:} 
Given $\mathcal{R}_{rank}$ in Eq.~\ref{eq:listrank}, the next step is to create a histogram $\mathcal{H}$. This step consists in analyzing which classifiers occur more frequently at top positions of $\mathcal{R}_{rank}$. For a better understanding, a fictitious example has been created in Tables~\ref{examplelisterror} and~\ref{occurhisto}. Table~\ref{examplelisterror} shows $\mathcal{R}_{rank}$ with a pool of classifiers for $\mathcal{C}=6$ classifiers sorted by $score_{rank} (c_{i}, c_{j})$, for a total of $15$ rows. Notice that $\mathcal{R}_{rank}$ is analyzed until all classifiers present in the pool occur at least once in the histogram $\mathcal{H}$. Table~\ref{occurhisto} shows the count of the classifiers until row~$8$, where the last classifier $c_3$ is added to the histogram $\mathcal{H}$.

\item \textbf{Roulette-Wheel Selection:} 
Given the histogram $\mathcal{H}$, we compute the probability of selecting each base classifier for composing the final ensemble through of a well-known literature technique called roulette. The probabilities are defined based on the frequency of each classifier in $\mathcal{H}$.

\item \textbf{Creating Ensembles:} 

The construction of each candidate ensemble is performed in two stages: (1) the random choice of the number of classifiers present in each ensemble, i.e. the number of times the roulette is played; and (2) the selection of the base classifiers through the probability roulette. Once a classifier is selected, it is added to the current ensemble.
\end{enumerate}

\begin{table}[!ht]
\centering
\caption{The final ranked list ($\mathcal{R}_{rank}$) ordered by $\textit{score}_{rank} (c_{i}, c_{j})$. Rows $1$-$8$ are the most suitable pairs of classifiers.}
\label{examplelisterror}

\resizebox{0.5\textwidth}{!}{
\begin{tabular}{lccc}
\toprule
Row & $c_i$ & $c_j$ & $score_{rank} (c_{i}, c_{j})$ \\ \hline
\rowcolor[HTML]{CCCCCC} 
 1 & $c_2$ & $c_4$ & 0.064 \\
\rowcolor[HTML]{CCCCCC} 
 2 & $c_4$ & $c_5$ & 0.169 \\
\rowcolor[HTML]{CCCCCC} 
 3 & $c_1$ & $c_5$ & 0.210 \\
\rowcolor[HTML]{CCCCCC} 
 4 & $c_2$ & $c_6$ & 0.265 \\
\rowcolor[HTML]{CCCCCC} 
 5 & $c_2$ & $c_5$ & 0.308 \\
\rowcolor[HTML]{CCCCCC} 
 6 & $c_1$ & $c_6$ & 0.400 \\
\rowcolor[HTML]{CCCCCC} 
 7 & $c_1$ & $c_2$ & 0.487 \\
\rowcolor[HTML]{CCCCCC}
 8 & $c_3$ & $c_6$ & 0.519 \\
\midrule
9 & $c_4$ & $c_6$ & 0.634 \\
10 & $c_3$ & $c_5$ & 0.636 \\
11 & $c_5$ & $c_6$ & 0.643 \\
12 & $c_1$ & $c_4$ & 0.673 \\
13 & $c_2$ & $c_3$ & 0.693 \\
14 & $c_1$ & $c_3$ & 0.821 \\
15 & $c_3$ & $c_4$ & 0.872 \\ \bottomrule
\end{tabular}
}
\end{table}

\begin{table}[!ht]
\centering
\caption{Histogram $\mathcal{H}$ created from ${R}_{rank}$.}
\label{occurhisto}
\resizebox{0.5\textwidth}{!}{
\begin{tabular}{@{}lcccccc@{}}
\toprule
Row & $c_1$ & $c_2$ & $c_3$ & $c_4$ & $c_5$ & $c_6$ \\ \midrule
1 &  & \textbullet &  & \textbullet &  &  \\
2 &  &  &  & \textbullet & \textbullet &  \\
3 & \textbullet &  &  &  & \textbullet &  \\
4 &  & \textbullet &  &  &  & \textbullet \\
5 &  & \textbullet &  &  & \textbullet &  \\
6 & \textbullet &  &  &  &  & \textbullet \\
7 & \textbullet & \textbullet &  &  &  &  \\
8 &  &  & \bet{\textbullet} &  &  & \textbullet \\ 
\hline
Frequency & 3 & 4 & 1 & 2 & 3 & 3 \\ \bottomrule
\end{tabular}
}
\end{table}
\subsection{Fitness Functions (F)}

The evaluation of the individuals in evolutionary algorithms is determined by the fitness function. This work investigated three fitness functions.

Let $e_m=1-accuracy(\mathcal{C}^*)$ be the mean error rate, and $\mathcal{C}^*$ the set of base classifiers existing in an ensemble. Equation~\ref{eq:error} shows the first approach, the fitness function $F_{E}$, which considers only $e_m$ to evaluate the ensembles:

\begin{equation}
    F_{E} = e_m.
    \label{eq:error}
\end{equation}

Equation~\ref{eq:diversity} shows the second approach, the fitness function $F_{D}$, which considers $e_m$ and the mean diversity scores ($d_m$) to evaluate the individuals. The score $d_m$ is defined as:

\begin{equation}
    d_m = \frac{\sum_{i=1}^{|C^*|}\sum_{j=i+1}^{|\mathcal{C}^*|} d_c(c_{ij})}{|\mathcal{C}^*|},
\end{equation}
\noindent
where $d_c(c_{ij})$ is mean diversity scores of the pair of classifiers $c_{ij} \in \mathcal{C}^{*}$. The function $F_{D}$ may be defined as follows:

\begin{equation}
    F_{D} = \frac{e_m + d_m}{2},
\label{eq:diversity}
\end{equation}

\noindent where $e_m, d_m \in [0,1]$. Finally, the proposed fitness function $F_{P}$ considers the two previous measures ($e_m$ and $d_m$) and pruning factor $t_p$ related to the amount of base classifiers $|\mathcal{C}^*|$ existing in the ensemble as:

\begin{equation}
\label{eq:fp}
    t_p = \frac{|\mathcal{C}^{*}|}{|\mathcal{C}|}.
\end{equation}

\noindent

Therefore, the fitness function $F_{P}$ is defined as:

\begin{equation}
    F_{P} = (\alpha \times e_m) + (\beta  \times d_m) + (\gamma \times t_p),
    \label{eq:pruning}
\end{equation}

\noindent where $\alpha$, $\beta$, and $\gamma \in [0,1]$ (with $\alpha+\beta+\gamma=1$)  are weighting coefficients for $e_m$, $d_m$, and $t_p$, respectively.

\subsection{Evolutionary Algorithms (E)}

In this step, the evolutionary algorithms (e.g., GA and UMDA) are used to search for optimal solutions by evolving the candidate ensembles. 
\subsection{Final Ensemble and Classification}

The best ensemble from the population optimized by GA and UMDA, i.e., the one that achieved the best fitness function score in the validation set,
is selected to classify the test set. The final classification for each instance on the test set is determined by a majority voting considering all $|\mathcal{C}^{*}|$ base classifiers of the ensemble.

\section{Experimental Methodology}
\label{sec:methodology}

\subsection{Datasets} \label{sec:datasets}

We have applied the proposed approach to 19 datasets collected from the UCI machine learning repository~\cite{uci}. Table~\ref{tab:datasets} describes each dataset in terms of the number of instances, attributes, and classes.

\begin{table}[ht!]
\centering
\caption{UCI datasets used in this work.}
\label{tab:datasets}
\centering
\resizebox{.7\textwidth}{!}{
\begin{tabular}{lccc}    \hline
\multicolumn{4}{c}{\textbf{UCI}}\\ \hline
\textbf{Datasets}  & \textbf{Instances}  & \textbf{Attributes} &  \textbf{Classes}\\ \hline
Balance-scale & 625 & 4 & 3\\ 
CMC & 1473 & 9 & 3\\
Ecoli & 336 & 8 & 8\\
Ionosphere & 351 & 34 & 2\\
Musk & 476 & 166 & 2\\ 
Page-blocks & 5473 & 10 & 5\\
Parkinsons & 195 & 22 & 2\\ 
Pen-digits & 10992 & 16 & 10\\
Phoneme & 5404 & 5 & 2\\ 
Pima & 768 & 8 & 2\\ 
Satimage & 6435 & 36 & 6\\
Segment & 2310 & 19 & 7\\
Spambase & 4601 & 57 & 2\\
Transfusion & 748 & 4 & 2\\
Wall-following & 5456 & 24 & 4\\
Waveform & 5000 & 21 & 3\\
Wine & 178 & 13 & 3\\
Wineq-red & 1599 & 11 & 6\\
Wineq-white & 4898 & 11 & 7\\ \hline
\end{tabular} 
}
\end{table}


\subsection{Diversity Measures}

Diversity is the degree of agreement/disagreement among involved classifiers 
pointing out the most interesting ones to be further used in a combination scheme. 
To achieve this diversity score inside ensemble systems, diversity measures have been used.
In~\cite{kuncheva03,Kuncheva_2004}, Kuncheva et al. compared several diversity measures, 
considering pairs of classifiers.

Let $\mathcal{M}$ be a matrix containing the relationship among a pair of classifiers with percentage of concordance 
that compute the percentage of \textit{hit} and \textit{miss} for two exemplifying classifiers $c_i$ and $c_j$.
The value $a$ is the percentage of examples that both classifiers $c_i$ and $c_j$ classified correctly in a validation set.
Values $b$ and $c$ are the percentage of examples that $c_j$ hit and
 $c_i$ missed and vice-versa. The value $d$ is the percentage of examples that both classifiers missed. 
   
 \begin{table} [htp!]%
 \centering
 \caption{Relationship matrix $\mathcal{M}$ among pairs of classifiers $c_i$ and $c_j$.}
 \begin{tabular}{|l|c|c|} \hline
 & Hit $c_i$ & Miss $c_i$ \\ \hline \hline
 Hit $c_j$&  $a$            & $b$ \\ \hline
 Miss   $c_j$&$c$            & $d$ \\ \hline
 \end{tabular}
 \label{tab:diversidade}
 \end{table}

In this work, five different measures have used \textit{Correlation Coefficient $\rho$} ($COR$),
\textit{Double-Fault Measure} ($DFM$), \textit{Disagreement Measure} ($DM$), 
 \textit{Interrater Agreement $k$} ($IA$),  and \textit{Q-Statistic} ($QSTAT$). 
 Those measures are defined as follows.
\begin{equation}
\label{eq:cor}
COR(c_i,c_j) =\frac{ad-bc}{\sqrt{(a+b)(c+d)(a+c)(b+d)}},
\end{equation}
\begin{equation}
\label{eq:dfm}
DFM(c_i,c_j) =d,  
\end{equation}
\begin{equation}
\label{eq:dm}
DM(c_i,c_j) =\frac{b+c}{a+b+c+d}.  
\end{equation}
\begin{equation}
\label{eq:ira}
IA(c_i,c_j) =\frac{2(ac-bd)}{(a+b)(c+d)+(a+c)(b+d)},  
\end{equation}

\begin{equation}
\label{eq:qstat}
QSTAT(c_i,c_j) =\frac{ad-bc}{ad+bc},  
\end{equation}

Diversity is higher if the measures \textit{Correlation Coefficient $p$}, 
\textit{Double-Fault Measure}, \textit{Interrater Agreement $k$}, and \textit{Q-Statistic} 
are lower between a pair of classifiers $c_i$ and $c_j$. In the case of the \textit{Disagreement Measure},
 the higher the measure, the greater the diversity~\cite{kuncheva03,Kuncheva_2004}. 
  

\subsection{State-of-the-art Approaches}
We have compared our approach with the following state-of-the-art ensemble selection techniques existing in the literature:  Aggregation Ordering in Bagging (AGOB)~\cite{AGOB_2004}, Diversity Regularized Ensemble Pruning (DREP)~\cite{DREP_2012}, Combining Diversity Measures for Ensemble Pruning (DivP)~\cite{DIVP_2016}, Genetic Algorithm based Selective ENsemble (GASEN)~\cite{GASEN_2002},
Kappa Pruning~\cite{kappa_pruning_1997}, and Pruning in Ordered Bagging Ensemble (POBE)~\cite{POBE_2006}.

\subsection{Evaluation Protocol}

We have used the same evaluation protocol adopted by state-of-the-art approaches in~\cite{DIVP_2016}: a 6-fold cross-validation protocol, where three partitions were selected to compose the training set, one partition to validation $1$, one partition to validation $2$, and one partition to test set. The final effectiveness result has been computed with the average accuracy of the six runs. 

\subsection{$CIF$-$E$ Protocol Configuration}\label{protocoloCIF-E}
 Table~\ref{tab:definicaoTecnicas} describes the four-step protocol \textbf{$CIF-E$}. For the $C$ parameter, two different modes can be chosen: one single type of technique (P) or multiple techniques (M). Therefore, the Perceptron technique was adopted as single technique (P) and for multiple techniques: the KNN with $k=\{1,3,5,7, 9,13,21\}$, {Decision Tree}, {Multilayer Perceptron}, {Gaussian Naive Bayes}, and {Perceptron}. {All learning techniques have used default parameters.}

\begin{table*}[ht!]
\centering
\caption{The $CIF-E$ protocol resulting in 24  different approaches ({C} $\times$ {I} $\times$ {F} $\times$ {E} = $2 \times 2 \times 3 \times 2$). }
\label{tab:definicaoTecnicas}
\resizebox{1\textwidth}{!}{%
\begin{tabular}{clcllcl}
\hline
\textbf{Protocol} &  & \multicolumn{2}{c}{\textbf{Parameter}} &  & \multicolumn{2}{c}{\textbf{Options}} \\ \cline{1-1} \cline{3-4} \cline{6-7} 
\multirow{10}{*}{$CIF-E$} &  & \textbf{Letter} & \multicolumn{1}{c}{\textbf{Meaning}} &  & \textbf{Mode} & \multicolumn{1}{c}{\textbf{Meaning}} \\ \cline{3-4} \cline{6-7} 
 &  & \multirow{2}{*}{C} & \multirow{2}{*}{Classifier} &  & P & Perceptron \\
 &  &  &  &  & M & Multiples techniques \\  \cline{2-7}
 &  & \multirow{2}{*}{I} & \multirow{2}{*}{Initialization} &  & A & Random \\
 &  &  &  &  & T & \textit{tuning}\\ \cline{2-7}
 &  & \multirow{3}{*}{F} & \multirow{3}{*}{Fitness Function} &  & E & Error \\
 &  &  &  &  & D & Error $+$ Diversity \\
 &  &  &  &  & P & Error $+$ Diversity $+$ Pruning \\ \cline{2-7}
 &  & \multirow{2}{*}{E} & \multirow{2}{*}{Evolutionary Algorithm} &  & GA & Genetic Algorithm \\
 &  &  &  &  & UMDA & {Univariate Marginal Distribuition Algorithm} \\ \hline
\end{tabular}%
}
\end{table*}

 The $I$ parameter is responsible for initializing the individual population of the evolutionary algorithm. Two modes were implemented, (A) for a random initialization; and (T) for \textit{tuning} -- both processes were described in Section ~\ref{sec:tuning}. The $F$ parameter has three different fitness functions to be used such as $F_{E}$ (error), $F_{D}$ (error $+$ diversity), and $F_{P}$ (error $+$ diversity $+$ number of classifiers). In the $F_{P}$, the ensembles with large number of base classifiers are penalized by the $\gamma$ coefficient. In this work, experiments using different values for $\gamma = \{0.1, 0.3,0.5\} $ were conducted and $\gamma=0.1$  achieved the best results. Finally, in the $E$ parameter, two alternatives of evolutionary algorithms are available to evolve the candidate ensembles: genetic algorithm (GA) and Univariate Marginal Distribution Algorithm (UMDA). Table~\ref{tab:evolutionary} shows the setup of each evolutionary algorithms.

\begin{table}[ht!]
    \centering
    \label{tab:evolutionary}
    \caption{Setup of each evolutionary algorithm (GA and UMDA) used in this work. Stagnation means the stopping criteria in case of non-evolution, i.e., the amount of generation with no progress of the best individual in the population. }
    \begin{tabular}{ccc} \toprule
       \textbf{Algorithm} & \textbf{Parameter} & \textbf{Value} \\ \midrule
        \multirow{6}{*}{GA} & Population size (P) & 500  \\
         & Number of generations ($G$) & 250  \\
          & Generation stagnation & 20\%  \\
           & Elitism (${e}$) & 40\%  \\
            & Mutation (${m}$) & 5\%  \\
             & Crossover (${c}$) & 30\%  \\ \hline
        \multirow{6}{*}{UMDA} & Population size ($\mu$) & 500  \\
         & Number of generations ($t$)& 250  \\
          & Generation stagnation & 20\%  \\
           & Initial probabilities ($p_t$) & 50\%  \\
            & Upper-bound margin ($1 - \frac{1}{n}$) & 95\%  \\
             & Lower-bound margin ($\frac{1}{n}$)& 5\%  \\ \bottomrule   
             \end{tabular}
    
    \label{tab:evolutionary}
\end{table}

\section{Results and Discussions}
\label{sec:experiments}

Three different experiments were performed. The first one shows an analysis between the different sizes of the available base classifier sets such as $|\mathcal{C}|=\{50,100,150,200,250\}$ that obtained the best accuracy results in the datasets. In the second, approaches based on the $CIF-E$ protocol were compared considering each of the four parameters that make up the proposed framework. Finally, in the third and last experiment, the best $CIF-E$ approach was compared with seven state-of-the-art methods existing in the literature.

\subsection{Choosing the size of the pool available classifier $|\mathcal{C}|$}

In this experiment, all twenty-four approaches implemented in the proposed $CIF-E$ protocol were grouped by the size of the available base classifier set ($|\mathcal{C}| = \{50,100,150,200,250 \}$). The median of these groups was calculated for each of the different datasets used in this work. 

As can be seen in Table~\ref{table:conjuntoDadosPool}, the pool of sizes $150$ and $200$ achieved similar results in the number of wins (five) considering the $19$ datasets analyzed, but the approaches using the $200$ had one more tie than the one of size $150$. We have therefore adopted $|C|=150$ for the analyzes of approaches in Section~\ref{sec:avaliacaoabordagensprotocoloCIF-E}.

\begin{table}[!ht]
\centering
\caption{Median accuracy of grouping all twenty-four $CIF-E$ protocol approaches for each size   $|\mathcal{C}|$. Highlights are the best accuracies per pool of classifiers.}

\label{table:conjuntoDadosPool}
\resizebox{0.7\textwidth}{!}{
\begin{tabular}{@{}llllll@{}}
\toprule
\multicolumn{1}{c}{\multirow{2}{*}{\textbf{Dataset}}} & \multicolumn{5}{c}{$|\mathcal{C}|$} \\ \cmidrule(l){2-6} 
\multicolumn{1}{c}{} & \textbf{50} & \textbf{100} & \textbf{150} & \textbf{200} & \textbf{250} \\ \midrule
Balance-scale & \bet{89.95} & 89.90 & 89.42 & 89.69 & 89.93 \\
CMC & 51.53 & 52.19 & 52.60 & \bet{52.80} & 52.04 \\
Ecoli & 69.64 & \bet{70.54} & 70.09 & \bet{70.54} & 69.64 \\
Ionosphere & 89.83 & 89.83 & 89.83 & 90.64 & \bet{90.68} \\
Musk & 80.13 & 81.01 & 80.82 & \bet{81.75} & 81.14 \\
Page-blocks & \bet{95.83} & 95.78 & 95.67 & 95.78 & 95.78 \\
Parkinsons & 81.53 & 81.53 & 83.07 & \bet{83.86} & 83.76 \\
Pendigits & 95.36 & 95.45 & 95.39 & 95.48 & \bet{95.66} \\
Phoneme & 81.32 & 81.74 & 81.94 & 82.35 & \bet{82.43} \\
Pima & 71.29 & 70.90 & 70.70 & 70.51 & \bet{71.29} \\
Satimage & 82.41 & 82.87 & \bet{83.83} & 83.22 & 83.64 \\
Segment & 92.86 & \bet{93.12} & 92.93 & 92.93 & 92.93 \\
Spambase & 90.28 & 90.19 & \bet{90.55} & 90.51 & 90.32 \\
Transfusion & 77.80 & \bet{77.96} & 77.91 & 77.71 & 77.51 \\
Wallfollowing & 81.74 & 81.67 & \bet{82.60} & 81.83 & 81.08 \\
Waveform & 85.12 & 85.57 & \bet{85.63} & 85.59 & 85.57 \\
Wine & 90.78 & 90.72 & 90.00 & \bet{90.78} & 89.83 \\
Wineq-red & 56.48 & 56.61 & \bet{58.07} & 58.07 & 57.97 \\
Wineq-white & 50.75 & 51.24 & 50.74 & \bet{52.04} & 51.37 \\ \midrule
\textbf{Win/Tie/Lose} & 2/0/17 & 2/1/16 & 5/0/14 & 5/1/13 & 4/0/15 \\ \bottomrule
\end{tabular}
}
\end{table}

\subsection{Evaluation of the $CIF-E$ protocol approaches}\label{sec:avaliacaoabordagensprotocoloCIF-E}

In order to define the best approach among the twenty-four implemented ones, four different experiments are performed for each parameter of the $CIF-E$ protocol.

\subsubsection{Evaluation of the parameter $C$}

In this experiments with the $C$ parameter of the $CIF-E$ protocol, an analysis of the twenty-four implemented approaches was performed, summing the number of wins and ties of each approach, according to the learning technique adopted (P and M).

\begin{figure}[!ht]
    \centering
    \includegraphics[width=1\textwidth]{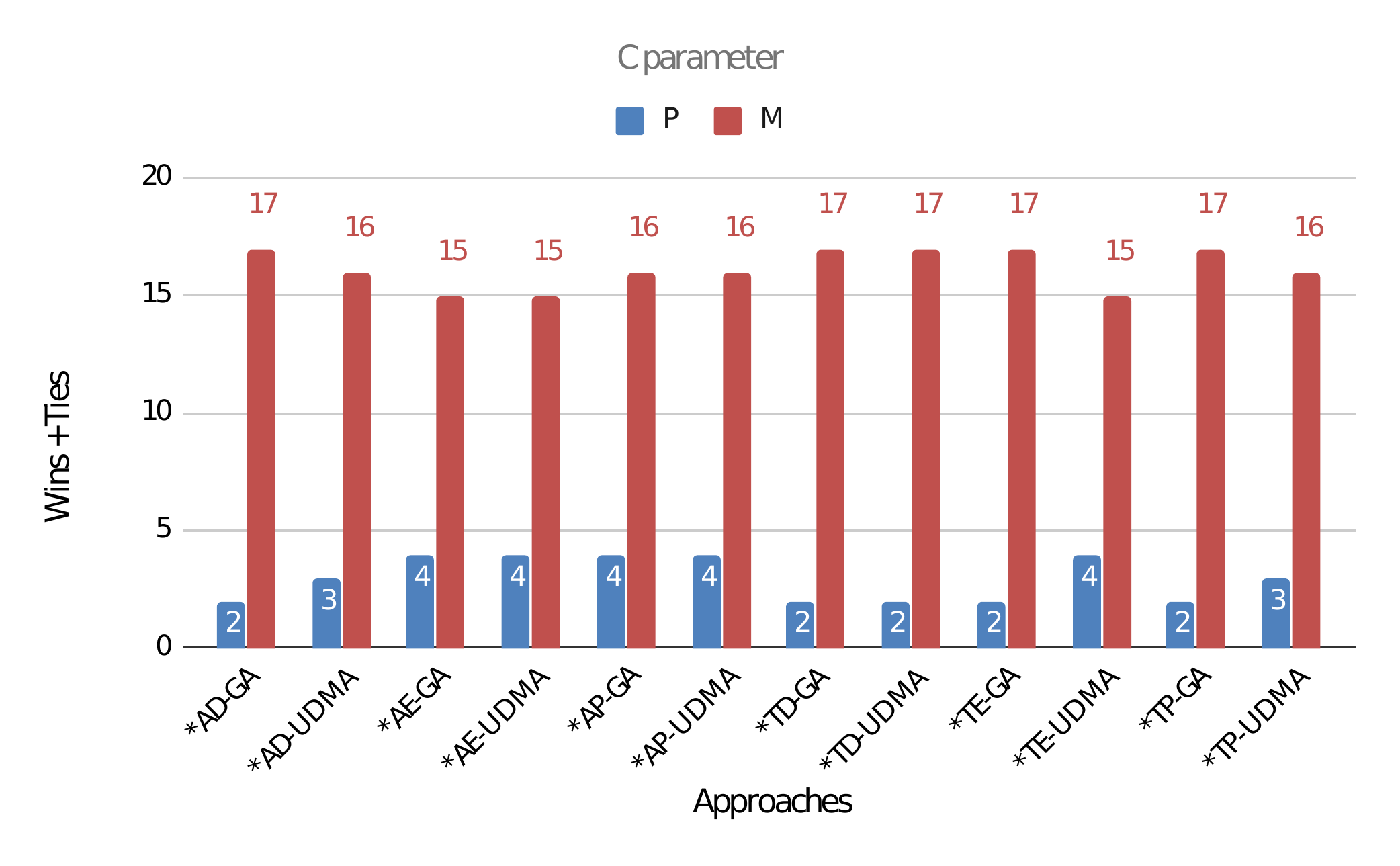}
    \caption{Number of wins $+$ ties among all of twenty four approaches grouped by type of learning technique (perceptron - P and multiple techniques - M) adopted in this experiment. }
    \label{fig:chartGeralParametroC}
\end{figure}

\begin{figure*}[!ht]
    \centering
    \includegraphics[width=1\textwidth]{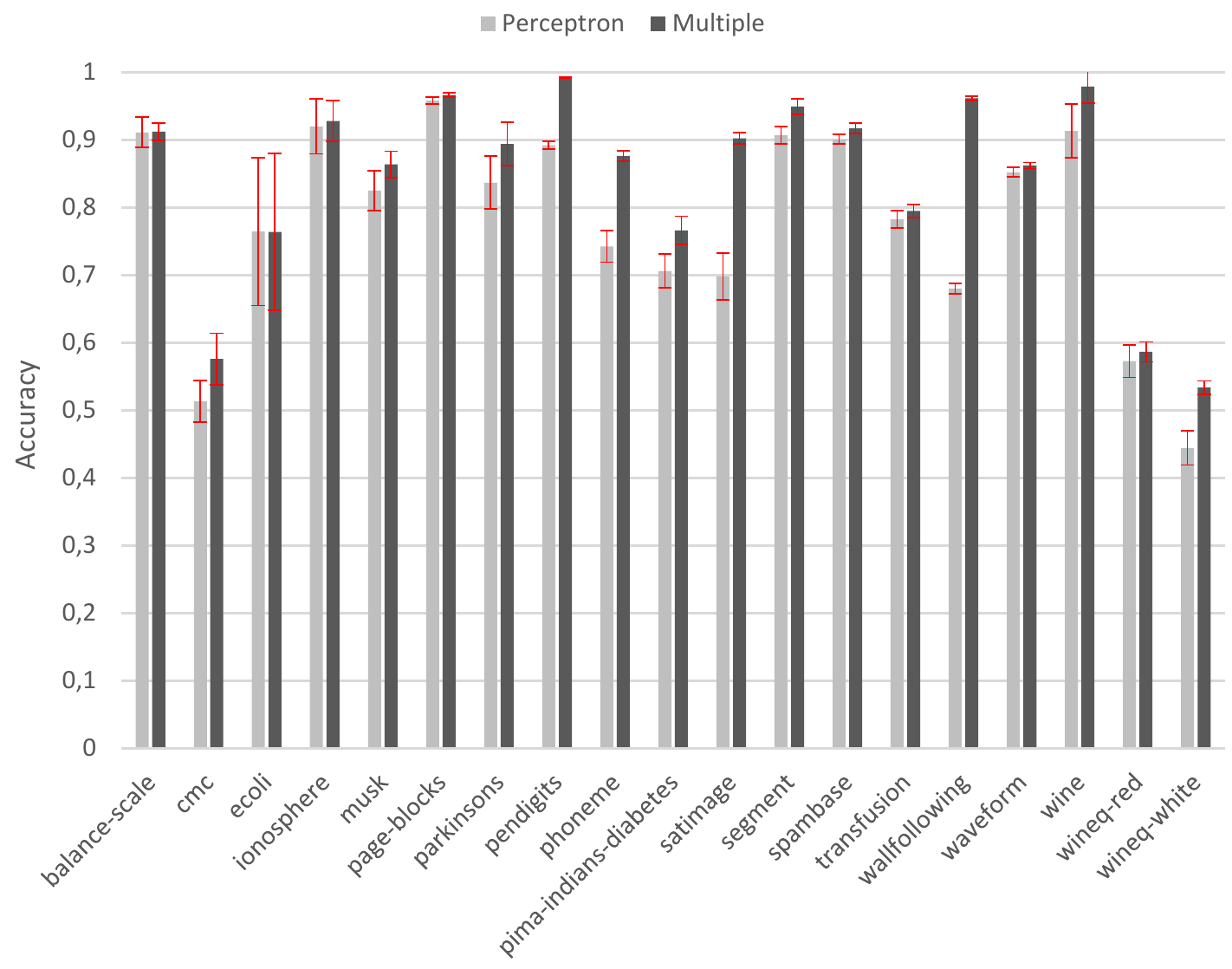}
    \caption{Mean accuracy and standard deviation (in red) results  of the base classifiers based on Perceptron (P) and multiple techniques (M). }
    \label{fig:EficienciaClassificadoresPerceptronHibrido}
\end{figure*}

In Figure~\ref{fig:chartGeralParametroC}, one may observe that approaches with multiple learning techniques (M) achieved better results than the ones with only the Perceptron technique (P). Moreover, an important factor to be analyzed is the accuracy achieved by the base classifiers. As can be seen in Figure~\ref{fig:EficienciaClassificadoresPerceptronHibrido}, twelve out of the nineteen datasets have similar results when we observe the standard deviations (in red) of both learning techniques. 

\subsubsection{Evaluation of the parameter $I$}

\begin{figure}[!ht]
    \centering
    \includegraphics[width=1\textwidth]{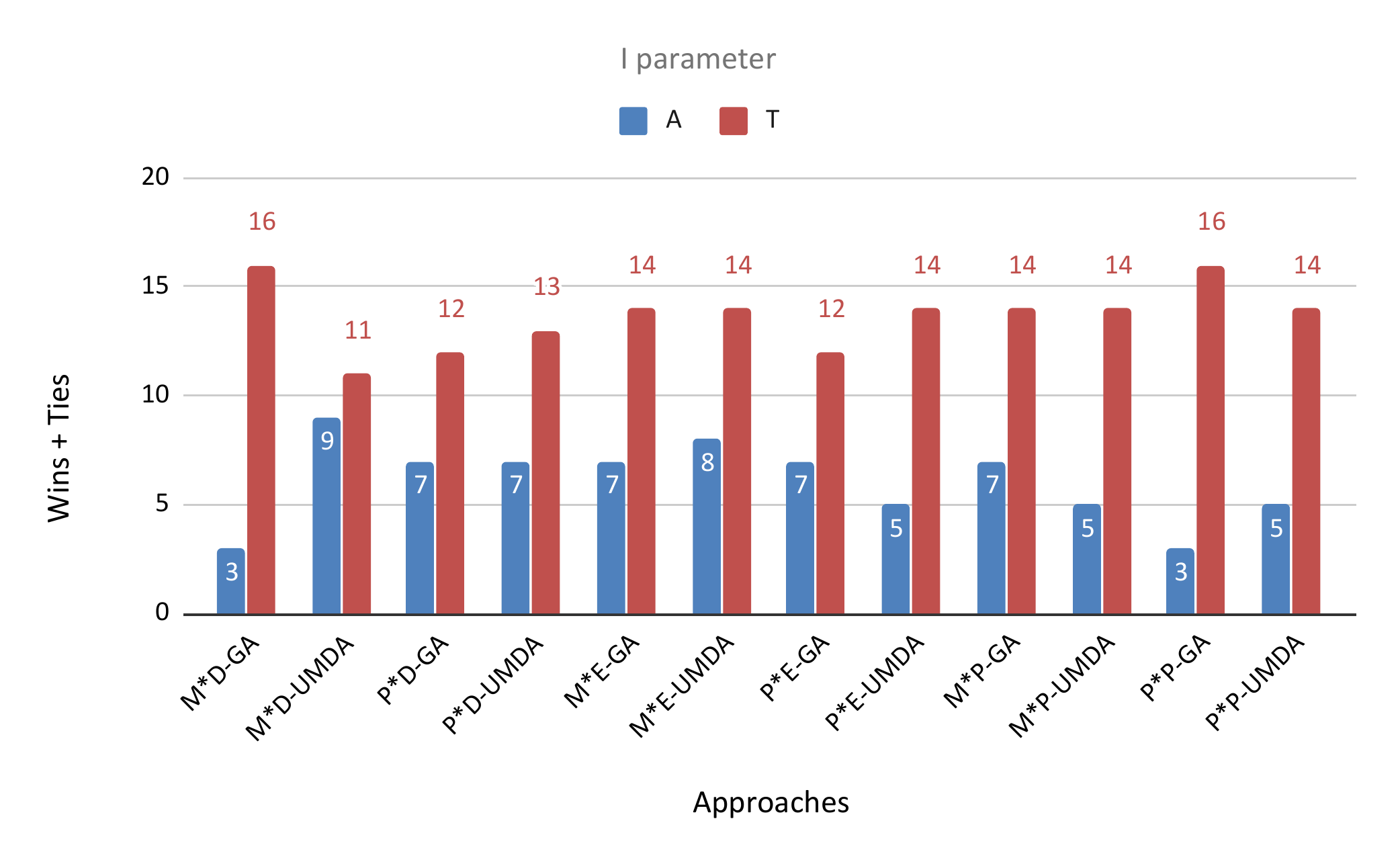}
    \caption{Number of wins $+$ ties among all of twenty four approaches grouped by type of initialization strategy (Random - A and Tuning - T) adopted in this experiment.}
    \label{fig:chartGeralParametroI}
\end{figure}

In the experiment for the $I$ parameter, two different initialization strategies can be employed to create the individuals (ensembles). This analysis compares the twelve approaches using random initialization (A) against the twelve approaches that use the {tuning} (T) initialization. Figure ~\ref{fig:chartGeralParametroI} shows the number of wins and ties achieved by each pair of equivalent approaches  in the $CIF-E$ protocol. For instance, {M*D-GA} means the MAD-GA and MTD-GA approaches use the same settings in the $CIF-E$ protocol, but they differ in the $I$ parameter. Therefore, it is possible observe that all approaches that use the \textit{tuning} (T) initialization achieved better results when compared to approaches that use random initialization (A).

\subsubsection{Evaluation of the $F$ parameter}

\begin{figure}[!ht]
    \centering
    \includegraphics[width=1\textwidth]{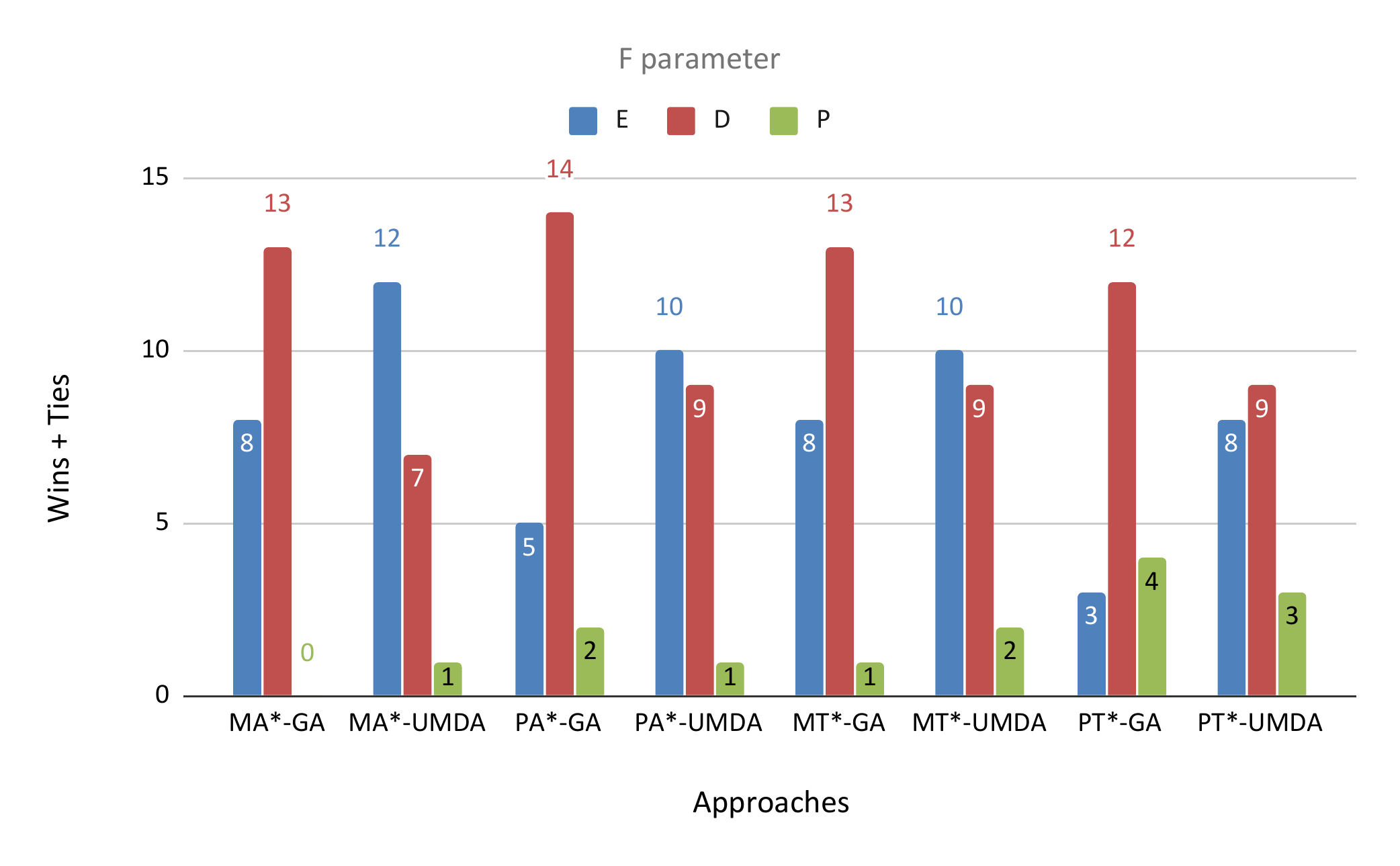}
    \caption{Number of wins $+$ ties among all of twenty four approaches grouped by type of fitness function (error - $F_{E}$, error$+$diversity - $F_{D}$, and error$+$diversity$+$number of classifiers - $F_{P}$) adopted in this experiment.}
    \label{fig:chartGeralParametroF}
\end{figure}

In the experiment with $F$ parameter, three different fitness functions ($F_{E}$, $F_{D}$, and $F_{P}$) can be used to evaluate the individuals/ensembles. 
Figure~\ref{fig:chartGeralParametroF} shows the number of wins and ties achieved by equivalent approaches  in the $CIF-E$ protocol. It is possible to observe that GA-based approaches (MAD-GA, PAD-GA, MTD-GA, and PTD-GA) achieved better results using $F_{D}$ fitness function. 
On the other hand, approaches based on UMDA (MAE-UMDA, PAE-UMDA, and MTE-UMDA) achieved better results using $F_{E}$ fitness function. Finally, $F_{P}$ fitness function did not achieve good results in any comparison. This issue may be justified by the penalisation that the individual receives for the amount of classifiers present in the ensemble. Therefore, the best results were obtained by individuals with high number of classifiers. The comparison of the amount of classifiers can be seen in the Table~\ref{tab:numeroClassificadoresProtocoloF}, where three approaches were compared using the same CIF-E protocol differentiating only fitness function (MTE-UMDA, MTD-UMDA, and MTP-UMDA). Notice that the average number of classifiers in the ensembles of the MTP-UMDA approach is much smaller (around $4$ base classifiers) than the average of the other two approaches (MTE-UMDA and MTD-UMDA), $37$ and $33$ respectively.

\begin{table}[!ht]
\centering
\caption{An analysis of the number of base classifiers present in the ensembles from each approach MT*-UMDA. i.e., three approaches using different fitness functions.}
\label{tab:numeroClassificadoresProtocoloF}
\resizebox{1\textwidth}{!}{%
\begin{tabular}{@{}lcccccccc@{}}
\toprule
                      & \multicolumn{2}{c}{\textbf{MTE-UMDA}} &  & \multicolumn{2}{c}{\textbf{MTD-UMDA}} &  & \multicolumn{2}{c}{\textbf{MTP-UMDA}} \\ \cmidrule(lr){2-3} \cmidrule(lr){5-6} \cmidrule(l){8-9} 
\textbf{Datasets} & \textbf{Accuracy}            & \#     &  & \textbf{Accuracy}            & \#     &  & \textbf{Accuracy}        & \#         \\ \midrule
balance-scale         & \bet{89.95}      & 12     &  & 89.42               & 13     &  & 89.47           & \betPruning{3} \\
cmc                   & \bet{55.39}      & 82     &  & 55.19               & 61     &  & 55.31           & \betPruning{4} \\
ecoli                 & 70.54               & 9      &  & \bet{71.43}      & 29     &  & \bet{71.43}  &    \betPruning{4} \\
ionosphere            & 91.53               & 27     &  & \bet{92.3}       & 31     &  & 92.24           & \betPruning{2} \\
musk                  & 84.38               & 49     &  & \bet{88.13}               & 40     &  & 85.44  & \betPruning{7} \\
page-blocks           & 96.71               & 42     &  & \bet{96.82}      & 27     &  & 96.05           & \betPruning{3} \\
parkinsons            & \bet{90.77}      & 15     &  & 89.25               & 32     &  & 84.61           & \betPruning{7} \\
pendigits             & \bet{99.26}      & 37     &  & 98.99               & 21     &  & 98.69           & \betPruning{3} \\
phoneme               & \bet{87.90}       & 42     &  & \bet{87.90}       & 36     &  & 87.06           &    \betPruning{3} \\
pima-indians-diabetes & 73.05               & 25     &  & 73.83               & 41     &  & \bet{75.39}  & \betPruning{6} \\
satimage              & 90.49               & 64     &  & \bet{90.72}      & 41     &  & 89.66           & \betPruning{2} \\
segment               & \bet{95.06}      & 43     &  & 94.94               & 39     &  & 94.03           & \betPruning{4} \\
spambase              & 93.29               & 35     &  & \bet{93.48}      & 27     &  & 92.70            & \betPruning{4} \\
transfusion           & \bet{78.32}      & 51     &  & 77.20                & 41     &  & 78.00              & \betPruning{6} \\
wallfollowing         & \bet{96.04}      & 23     &  & 95.55               & 15     &  & 93.73           & \betPruning{3} \\
waveform              & \bet{86.31}      & 52     &  & 86.08               & 33     &  & 84.63           & \betPruning{3} \\
wine                  & 93.22               & \betPruning{4}      &  & \bet{94.94}      & 25     &  & 93.33           & 7 \\
wineq-red             & 59.66               & 42     &  & \bet{61.73}      & 29     &  & 57.68           & \betPruning{4} \\
wineq-white           & 55.94               & 50     &  & \bet{56.34}      & 36     &  & 53.79           & \betPruning{4} \\ \bottomrule 

{\textbf{Average of Classifiers}}           &                & 37     &  &       & 33     &  &            & \betPruning{4} \\ \bottomrule 

\end{tabular}%
}
\end{table}

\subsubsection{Evaluation of the parameter $E$}

\begin{figure}[!ht]
    \centering
    \includegraphics[width=1\textwidth]{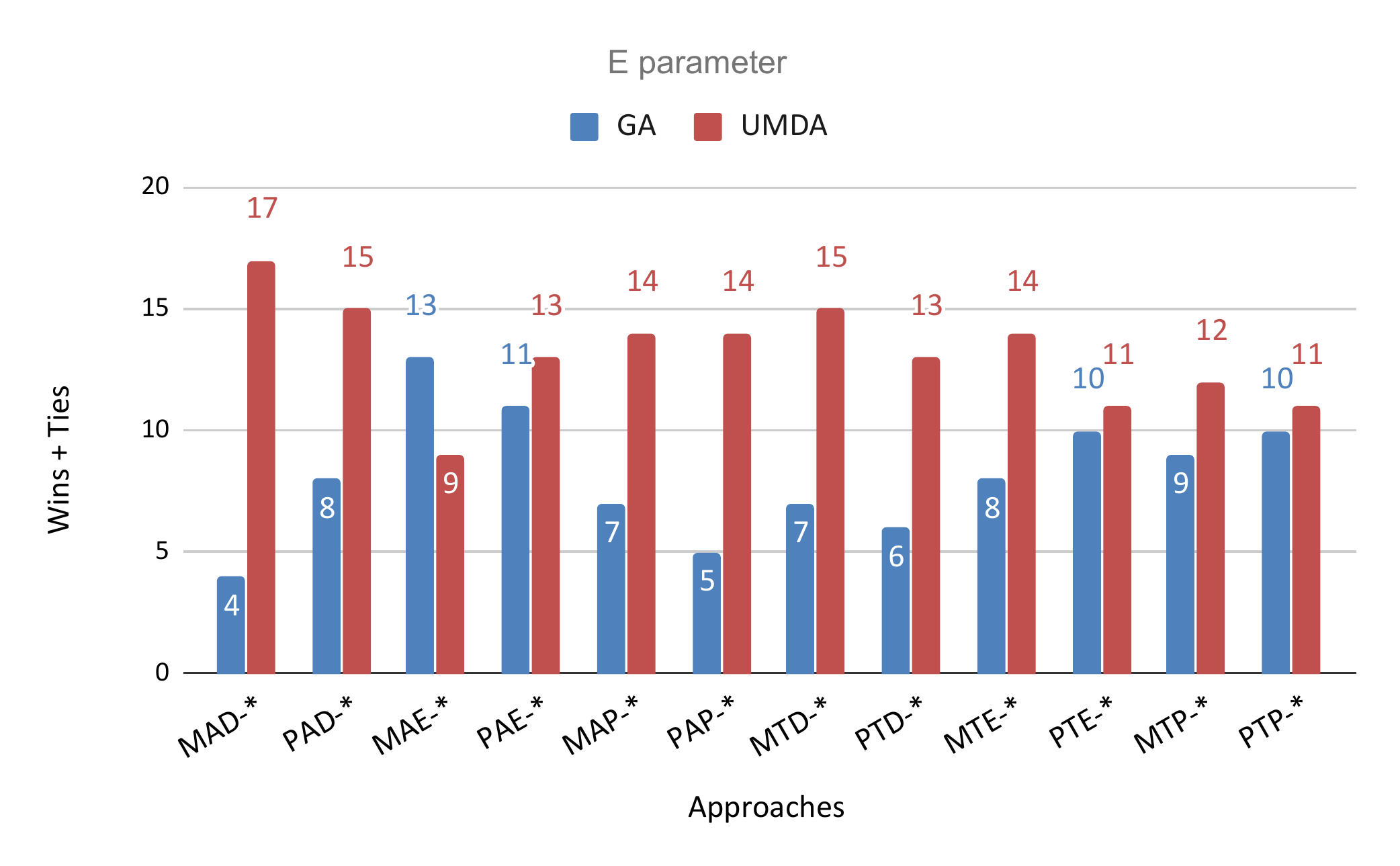}
    \caption{Number of wins $+$ ties among all of twenty four approaches grouped by type of evolutionary algorithm (genetic algorithm - GA and univariate   marginal   distribution   algorithm - UMDA) adopted in this experiment..}
    \label{fig:chartGeralParametroE}
\end{figure}

In the experiment with the parameter $F$, two different evolutionary algorithms were employed (GA and UMDA). Figure~\ref{fig:chartGeralParametroE} shows the number of wins and ties achieved by the equivalent approach in the $CIF-E$ protocol. It is possible to observe the huge advantage that UMDA-based approaches had over GA-based approaches, resulting in eleven (out of twelve) wins and a single lost (MAE-GA $\times$ MAE-UMDA).

\subsection{Defining the Best $CIF-E$ Approach}

Considering the previous experiments performed in Section~\ref{sec:avaliacaoabordagensprotocoloCIF-E}, it is possible to identify the best  parameter combination of the $CIF-E $ protocol to select the best approach among all of twenty four implemented here. The best approach has multiple techniques (M), \textit{tuning} initialization (T), $F_{D}$ and the UMDA algorithm, i.e., MTD-UMDA approach. In the next section, that best approach is compared to the best methods in the literature. In addition, for a fairer comparison, another approach is selected (PTP-UMDA) since the baselines  use only Perceptron techniques as base classifiers.

\subsection{Accuracy Comparison among State-of-the-art Methods}
\label{sec:comparacaoentremelhorabordagemCIF-Ebaseline}
Our MTD-UMDA approach was compared with seven other approaches (AGOB\cite{AGOB_2004}, Bagging\cite{bagging1996}, DivP\cite{DIVP_2016}, DREP\cite{DREP_2012}, GASEN\cite{GASEN_2002}, Kappa\cite{kappa_pruning_1997} and POBE\cite{POBE_2006}) from the literature. In Table ~\ref{tab:approachesMultiplos}, it can be observed that MTD-UMDA achieved better results in fourteen out of the nineteen UCI datasets. The best baseline was $DREP$, which achieved better results in two datasets. It is very important comment that the best $CIF-E$ approach using Perceptron (PTD-UMDA) also achieved two wins. Furthermore, in comparison with the number of classifiers used by ensembles from each approach, it is possible to note that the ensemble created by PTD-UMDA used slightly less base classifiers than the best baseline ($DivP$), achieved $8$ wins, $6$ losses, and $5$ ties with an average of classifiers equal to $3$. Finally, PTD-UMDA has much fewer base classifiers than other baselines, including the most accurate approach (MTD-UMDA). 

Table~\ref{tab:testEstatisticoMultiplos} shows a significance statistic test among all approaches compared in this experiment and it shows that our MTD-UMDA approach is statistically different from the competitors.

\subsection{Efficiency Analysis}
In this analysis, two approaches (MTD-UMDA and PTP-UMDA) were compared against each other in terms of computational efficiency. The adopted protocol was $6$-fold cross validation using $\mathcal{C}=150$ base classifiers as pool. All experiments of training and test were performed on a computer with AMD Ryzen 5/2600 processor and 16GB RAM. Table~\ref{tab:time} shows performance of the approaches in terms of "seconds" spent in this experiment for each UCI dataset.

As can be observed, PTP-UMDA is, on average, $5.55$ times faster than MTD-UMDA. This difference is related to the fitness function used by each approach implemented on Python/scikit-learn. It is very important comment that the same analysis was not performed among baseline approaches, since they have been implemented on Matlab and such analysis might be unfair. For instance, DivP was found to be $35$ times slower than PTP-UMDA using the same experimental protocol, but running on Matlab. 

\begin{table}[ht]
    \caption{Wilcoxon Signed-Rank Tests~\cite{Wilcoxon1992} are displayed using significance level at $5\%$.} 
    \label{tab:testEstatisticoMultiplos}
    \centering
    \begin{tabular}{lc} \\ \toprule
    \textbf{Approaches}     & \textbf{p-value} \\ \midrule
MTD-UMDA $ \times$ AGOB  & 0.0017 \\
MTD-UMDA $ \times$ Bagging & 0.0017 \\
MTD-UMDA $ \times$ DivP & 0.0043 \\
MTD-UMDA $ \times$ DREP & 0.0057 \\
MTD-UMDA $ \times$ GASEN & 0.0019 \\
MTD-UMDA $ \times$ Kappa & 0.0008 \\
MTD-UMDA $ \times$ POBE & 0.0011 \\
MTD-UMDA $ \times$ PTP-UMDA & 0.0003 \\\bottomrule
\end{tabular}
\end{table}

\begin{table}[ht]
\caption{Efficiency analysis among PTP-UMDA and MTD-UMDA approaches. Time in seconds.}
    \centering
    \resizebox{1\textwidth}{!}{
    \begin{tabular}{c|cc|c} \hline
\textbf{Datasets} & \textbf{PTP-UMDA} & \textbf{MTD-UMDA} & \textbf{MTD/PTP}\\ \hline
Balance-scale  & 68 & 234 & 3.44\\
CMC  & 101 & 1097 & 10.86\\
Ecoli  & 52 & 206 & 3.96\\
Ionosphere  & 48 & 191 & 3.98\\
Musk  & 59 & 302 & 5.12\\
Page-blocks  & 370 & 1249 & 3.38\\
Parkinsons  & 52 & 239 & 4.60\\
Pen-digits  & 603 & 2151 & 3.57\\
Phoneme  & 364 & 1684 & 4.63\\
Pima  & 72 & 503 & 6.99\\
Satimage  & 393 & 2864 & 7.29\\
Segment  & 169 & 961 & 5.69\\
Spambase  & 388 & 1564 & 4.03\\
Transfusion  & 82 & 533 & 6.50\\
Wall-following  & 428 & 1139 & 2.66\\
Waveform  & 349 & 2431 & 6.97\\
Wine  & 16 & 145 & 9.06\\
Wineq-red  & 78 & 527 & 6.76\\
Wineq-white  & 171 & 1011 & 5.91\\\hline
\textbf{Average} & {203.3} & 1001.6 & 5.55\\\hline
\end{tabular}
}\label{tab:time}
\end{table}

\section{Conclusions}
\label{sec:conclusion}
This work proposed a framework of classifier selection and fusion based on $CIF-E$ protocol, which uses evolutionary algorithms to find the best final ensemble for a target application.

In the performed experiments, five different analysis have been done.
First, five different sizes of the pool available classifier $|C| = \{50, 100, 150, 200, 250\}$ might show that the pools of $150$ and $200$ base classifiers achieve similar classification results and $|C|=200$ was adopted for next analysis. 
Second, twenty four approaches from $CIF-E$ protocol ($C \times I \times F \times E = 2 \times 2 \times 3 \times 2 = 24$) were implemented and validated. This analysis might show that approaches with multiple learning techniques (M) achieve better results than single techniques (P). Tuning initialization (T) shows to be better than random initialization (R). Fitness function that combines accuracy and diversity ($F_{D}$) is better than one that use only accuracy ($F_{E}$) or combine accuracy, diversity, and amount of classifiers ($F_{P}$). Finally, the UMDA-based approaches show better performance than GA-based approaches. 

Third, taking into account the previous analysis, the best found approach (MTD-UMDA) was compared to seven other approaches existing in the literature. In this analysis, MTD-UMDA achieved better results in fourteen datasets out of the nineteen used in this work. A statistical significance test might show that MTD-UMDA is better than all baselines.

Forth, a variation of the best approach (PTP-UMDA) using different learning technique and fitness function than MTD-UMDA was compared considering the number of base classifiers present in the final ensemble created by each approach. In this analysis, PTP-UMDA achieves to select slightly less base classifiers than the best pruning baseline (DivP). 

In last analysis, an efficiency experiment was performed among MTD-UMDA and PTP-UMDA. PTP-UMDA approach is, on average, 5.55 times faster than MTD-UMDA approach.

Finally, the proposed framework showed to be a generic and flexible solution, which allows to change its protocol steps $CIF-E$ for specific applications. Therefore, providing agility in the development and validation of new methods, reducing time and costs in development of solutions. As future work, some points can be tested, such as reducing the number of classifiers in ensembles while maintaining high accuracy (multi-objective task); individual assessment of each diversity measure and the impact on classifier selection; individuals' initialization of evolutionary algorithms; addition of new learning techniques to increase the diversity of the generated base classifiers and addition of new evolutionary algorithms for the evolution process of the candidate ensembles.

\section{ACKNOWLEDGMENTS}
The authors thank the support of the S\~{a}o Paulo Research Foundation (FAPESP) through grants \#2018/23908-1 and \#2017/25908-6. Also the Brazilian scientific funding agency CNPq through the Universal Project (grant \#408919/2016-7).

    \begin{landscape}{
        \begin{table}[]
        \centering

            \caption{Median of the accuracy results among different approaches existing in the literature using $|\mathcal{C}|=150$ available classifiers for nineteen UCI datasets. The best results are highlighted and \# means the number of base classifiers present in the ensemble.} 
            
            \label{tab:approachesMultiplos}
            \resizebox{1.6\textwidth}{!}{
                \begin{tabular}{lcccccccccccccccccc}
                \hline
                \textbf{Datasets} &  \textbf{AGOB}                            &             \#  &  \textbf{Bagging}                        &  \#   &  \textbf{DivP}                            &  \#  &  \textbf{DREP}                            &  \#  &  \textbf{GASEN}   &  \#   &  \textbf{Kappa}   &  \#  &  \textbf{POBE}    &  \#  &  \textbf{MTD-UMDA}                        &  \#  & \textbf{PTP-UMDA} &  \#  \\ \hline
Balance-scale          & 88.3 & 23 & 87.4 & 150 & 90.1 &  \bet{1}   & 88.3 & 75 & 87.2 & 78 & 88.2 & 30 & 89.6 & 27 & 89.0 & 16 & \bet{90.4} & 3 \\
CMC                    & 51.0 & 41 & 50.5 & 150 & 48.9 &  {5}   & 50.4 & 75 & 50.5 & 141 & 50.4 & 30 & 49.6 & 35 &  \bet{55.4}  & 45 & 49.4 &  \bet{2} \\
Ecoli                  &  \bet{79.7}                      & 28 & 79.4 & 150 & 78.3 &  \bet{2}   &  \bet{79.7}  & 75 & 79.4 & 75 & 78.2 & 30 & 78.5 & 31 & 70.5 & 18 & 71.4 & 3 \\
Ionosphere             & 88.6 & 33 & 88.3 & 150 &  {88.6}  &  \bet{2}   & 88.6 & 75 & 88.1 & 66 & 88.9 & 30 & 90.0 & 31 &  \bet{90.5}                            & 12 & 88.9 &  \bet{2}  \\ 
Musk                   & 78.4 & 27 & 78.6 & 150 & 79.2 &  \bet{2}   & 79.0 & 75 & 79.6 & 79 & 78.8 & 30 & 79.8 & 30 &  \bet{85.6}  & 26 & 77.9 & 3 \\
Page-blocks            & 95.5 & 24 & 95.0 & 150 & 95.4 & 3 & 95.7 & 75 & 95.0 & 141 & 94.8 & 30 & 95.0 & 40 &  \bet{96.8}  & 15 & 95.1 &  \bet{2}   \\
Parkinsons             & 83.1 & 26 & 82.6 & 150 & 84.6 &  \bet{1}   & 83.1 & 75 & 83.1 & 71 & 81.1 & 30 & 82.1 & 34 &  \bet{90.8}  & 24 & 80.0 & 5 \\
Pen-digits             & 90.7 & 23 & 90.6 & 150 & 90.8 & 10 & 91.1 & 75 & 90.6 & 143 & 91.2 & 30 & 90.9 & 35 &  \bet{99.0}  & 17 & 90.0 &  \bet{2}   \\
Phoneme                & 71.3 & 27 & 72.1 & 150 & 76.7 &  \bet{2}   & 72.5 & 75 & 71.3 & 141 & 66.8 & 30 & 71.9 & 32 &  \bet{87.8}  & 28 & 76.8 & 3 \\
Pima                   & 74.1 & 19 & 74.1 & 150 & 74.1 &  \bet{3}  &  \bet{75.9}  & 75 & 74.5 & 113 & 69.7 & 30 & 71.2 & 34 & 73.8 & 31 & 68.0 &  \bet{3}   \\
Satimage               & 64.2 & 36 & 64.2 & 150 & 66.9 & 4 & 64.3 & 75 & 64.3 & 142 & 63.5 & 30 & 65.1 & 38 &  \bet{90.6}  & 29 & 75.9 &  \bet{2}   \\
Segment                & 91.2 & 36 & 91.2 & 150 & 91.3 & 4 & 91.0 & 75 & 91.3 & 114 & 91.4 & 30 & 91.1 & 39 &  \bet{94.4}  & 30 & 90.8 &  \bet{2}   \\
Spambase               & 91.7 & 37 & 92.3 & 150 &  {92.7}  & 7 & 92.6 & 75 & 92.4 & 114 & 90.4 & 30 & 92.1 & 33 &  \bet{93.4}                            & 20 & 90.0 &  \bet{2}   \\
Transfusion            & 68.9 & 21 & 74.9 & 150 &  {77.1}  &  \bet{3}   & 77.1 & 75 & 74.7 & 74 & 65.0 & 30 & 74.7 & 34 &  {77.6}                            & 37 & \bet{78.3} &  \bet{3}   \\
Wall-following         & 64.6 & 40 & 65.7 & 150 & 66.5 & 10 & 65.8 & 75 & 65.8 & 143 & 61.5 & 30 & 66.4 & 36 &  \bet{95.3}  & 12 & 69.2 & \bet{4} \\
Waveform               & 82.9 & 30 & 85.2 & 150 &  {85.2}  &  \bet{3}   & 85.2 & 75 & 85.1 & 110 & 80.0 & 30 & 82.6 & 36 &  \bet{86.4}                            & 29 & 83.1 &  \bet{3}  \\ 
Wine                   & 95.5 & 19 & 96.0 & 150 &  \bet{97.2}  &  \bet{1}   &  {96.6}  & 75 & 96.0 & 74 & 96.1 & 30 & 95.5 & 22 &  {96.6}  & 19 & 88.3 & 4 \\
Wineq-red              & 56.0 & 41 & 56.3 & 150 & 57.2 & 6 & 57.0 & 75 & 56.5 & 142 & 54.5 & 30 & 54.5 & 35 &  \bet{63.1}  & 29 & 58.4 & \bet{3} \\
Wineq-white             & 50.2 & 32 & 48.6 & 150 & 50.2 &  \bet{2}   & 49.4 & 75 & 48.6 & 142 & 48.7 & 30 & 47.0 & 40 &  \bet{55.9}  & 25 & 50.5 &  \bet{2}  \\ \hline 
\textbf{Average of classifiers}                 &                                  & 30 &                                         & 150 &                                  & 4 &                                  & 75 &          & 111 &          & 30 &          & 34 &                                  & 24 &  & \bet{3}  \\
\textbf{Win/Tie/Loss}  &  0/1/18                          &      &  0/0/19                                 &       &  1/0/18                          &      & 1/1/17 &      &  0/0/19  &       &  0/0/19  &      &  0/0/19  &      &  14/0/5                          &      &  2/0/17    &  \\ \hline

                \end{tabular}
            }
        \end{table}
    }\end{landscape}
    
\bibliographystyle{model5-names}\biboptions{authoryear}
\bibliography{ref}

\end{document}